%% file: main.tex
\documentclass[conference]{IEEEtran}
\IEEEoverridecommandlockouts

\usepackage{paralist}
\usepackage{cite}
\usepackage{amsmath,amssymb,amsfonts}
\usepackage{algorithmic}
\usepackage{graphicx}
\usepackage{textcomp}
\def\BibTeX{{\rm B\kern-.05em{\sc i\kern-.025em b}\kern-.08em
    T\kern-.1667em\lower.7ex\hbox{E}\kern-.125emX}}

\usepackage[table]{xcolor}

\input{dfn}

\usepackage{graphicx}
\usepackage{subfigure}
\usepackage{booktabs} 
\usepackage{amssymb}
\usepackage{tabularx}
\usepackage{makecell}
\usepackage{hyperref}
\usepackage{amsfonts} 
\usepackage{amsmath} 
\usepackage{bbm}
\usepackage{array,multirow,graphicx}
\usepackage{bm}
\usepackage{etoolbox,xspace}
\usepackage{footmisc}
\usepackage{algorithm}
\usepackage{algorithmic}
\usepackage[most]{tcolorbox}

\begin{document}

\title{{\huge{\methodtitle: Anomaly Detection of Attributed Multi-graphs with Metadata: A Unified Neural Network Approach}}\\
\thanks{$^\ast$Equal contribution. Research supported by the Advanced Research Projects Activity (IARPA) via Department of Interior/Interior Business Center (DOI/IBC) contract number 140D0423C0033 and the PwC Digital Transformation and Innovation Center at Carnegie Mellon University Intelligence. The U.S. Government is authorized to reproduce and distribute reprints for Governmental purposes notwithstanding any copyright annotation thereon. Disclaimer: The views and conclusions contained herein are those of the authors and should not be interpreted as necessarily representing the official policies or endorsements, either expressed or implied, of IARPA, DOI/IBC, the U.S. Government, or the other funding parties. }
}
\author{
\IEEEauthorblockN{Konstantinos Sotiropoulos$^\ast$}
\IEEEauthorblockA{\textit{Heinz College}} 
\textit{Carnegie Mellon University}\\
{ksotirop@andrew.cmu.edu}
\and
\IEEEauthorblockN{Lingxiao Zhao$^\ast$}
\IEEEauthorblockA{\textit{Heinz College}} 
\textit{Carnegie Mellon University}\\
{lingxia1@andrew.cmu.edu}
\and
\IEEEauthorblockN{Pierre Jinghong Liang}
\IEEEauthorblockA{\textit{Tepper School of Business}} 
\textit{Carnegie Mellon University}\\
{liangj@andrew.cmu.edu}
\and
\IEEEauthorblockN{Leman Akoglu}
\IEEEauthorblockA{\textit{Heinz College}} 
\textit{Carnegie Mellon University}\\
{lakoglu@andrew.cmu.edu}
}

\maketitle

\begin{abstract}
Given a complex graph database of node- and edge-attributed multi-graphs as well as associated metadata for each graph, how can we spot the anomalous instances? Many real-world problems can be cast as graph inference tasks where the graph representation could capture complex relational phenomena (e.g., transactions among financial accounts in a journal entry), along with metadata reflecting tabular features (e.g. approver, effective date, etc.). 
While numerous anomaly detectors based on Graph Neural Networks (GNNs) have been proposed, none are capable of directly handling directed graphs with multi-edges and self-loops. Furthermore, the simultaneous handling of relational and tabular features remains an unexplored area.
In this work we propose \method, a novel graph neural network model that handles directed multi-graphs, providing a unified end-to-end architecture that fuses metadata and graph-level representation learning through an unsupervised anomaly detection objective. Experiments on datasets from two different domains, namely, general-ledger journal entries from different firms (accounting) as well as human GPS trajectories from thousands of individuals (urban mobility), 
validate \method's generality and detection effectiveness of expert-guided and ground-truth anomalies. Notably, \method  outperforms existing baselines that handle the two data modalities (graph and metadata)  separately with post hoc synthesis efforts.
\end{abstract}


\vspace{-0.05in}
\section{Introduction}
\label{sec:intro}

\input{01intro}

\section{Preliminaries}
\label{sec:prelim}

\input{02prelim}

\section{\method for Multi-modal Anomaly Detection of (Multi-)Graphs with Metadata}
\label{sec:measures}

\input{03method}

\section{Experiments}
\label{sec:experiment}

\input{04experiments}

\section{Related Work}
\label{sec:related}

\input{05related}

\section{Conclusion}
\label{sec:conclusion}

\input{06conclusion}

\bibliography{00refs.bib}
\bibliographystyle{IEEEtran}

\newpage
\input{07appendix}
\end{document}

%% file: dfn.tex
\newsavebox\CBox

\newcommand{\methodtitle}{{\LARGE \sf ADAMM}}
\newcommand{\method}{{\sf ADAMM}\xspace}

\newcommand{\cbit}{\begin{compactitem}}
	\newcommand{\ceit}{\end{compactitem}}
\newcommand{\cben}{\begin{compactenum}}
	\newcommand{\ceen}{\end{compactenum}}

\definecolor{OliveGreen}{rgb}{0,0.6,0}

\newcommand{\bit}{\begin{itemize}}
	\newcommand{\eit}{\end{itemize}}
\newcommand{\ben}{\begin{enumerate}}
	\newcommand{\een}{\end{enumerate}}
\newcommand{\beq}{\begin{equation}}
\newcommand{\eeq}{\end{equation}}

\newcommand{\bx}{\bold{x}}

\newcommand{\hide}[1]{}

\newtheorem{problem}{\textbf{Problem}}

\newtheorem{definition}{\textbf{Definition}}
\newcommand{\blf}{\bold{f}}
\newcommand{\blt}{\boldsymbol{\theta}}

\newcommand{\blC}{\bold{C}}

\newcommand{\blc}{\boldsymbol{c}}
\newcommand{\bZ}{\bold{Z}}
\newcommand{\blg}{\boldsymbol{\gamma}}
\newcommand{\blG}{\bold{\Gamma}}

\usepackage{array}
\newcolumntype{C}[1]{>{\centering\arraybackslash\hspace{0pt}}p{#1}}

%% file: 01intro.tex

Anomaly detection finds numerous practical applications in finance, manufacturing, monitoring, etc. as anomalies are typically indicators of faults, inefficiencies, malicious behavior, etc. in  various real-world systems. One of the key challenges in real world settings is the complexity of the data, which exhibit multiple different modalities and heterogeneity---requiring new data representations and novel modeling designs.

This work is motivated by anomaly detection problems in two different real-world domains. The first is from business management and particularly accounting/auditing, where the goal is to identify abnormalities (errors or fraud) among annual general-ledger journal entries from a given firm. 
Each entry consists of a series of line items of credit or debit transactions of various amounts between accounts with the total debited dollar amount equal to the total credited amount, following the double-entry bookkeeping rules.
Accordingly, debits and credits within an entry create directed and weighted links between accounts, and multiple transactions may occur between the same pair of accounts or even within a single general-ledger account.
Besides the relational information, each journal entry is also associated with meta-features, such as the approver, entry and effective dates, etc. 
This poses a multi-modal  (relational and tabular) data problem setting.
A second example arises from communication networks, where the problem is detecting significant events within a company based on e-mails exchanged between different entities.

\begin{figure}[!t]
    \centering
    \begin{tabularx}{\linewidth}{XX}
 \includegraphics[width=\linewidth] {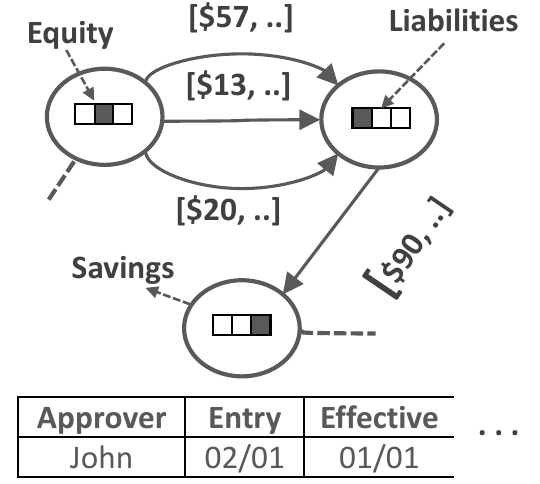}
&
\includegraphics[width=\linewidth] {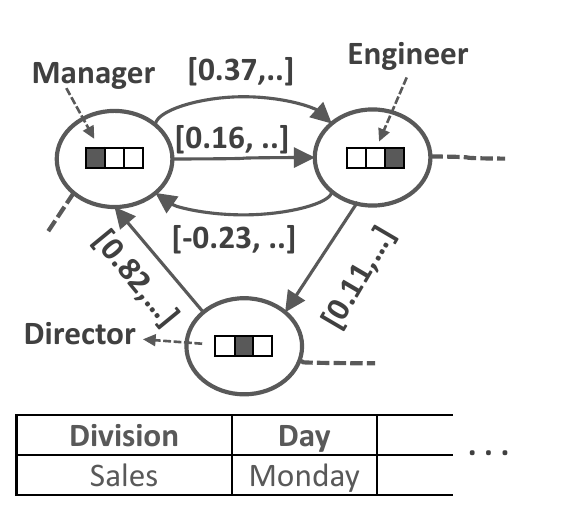} 
    \end{tabularx}
    \vspace{-2mm}
    \caption{
        Modeling complex data. {(left)} E.g. from \textbf{accounting}: a journal entry's attributed \textit{multi-graph} (multiple transactions between two accounts), with edge directions (credit/debit), edge features (e.g. \$ amount), and node features (account type; e.g. equity, savings, etc.) plus aux. \textit{meta-features} (approver, entry date, etc.); 
        {(right)} E.g. from \textbf{communication networks}: a daily activity \textit{multi-graph} (multiple e-mails between two company employees), with edge directions (to/from), edge features (e.g. text embedding) and node features (role in the company.) plus aux. \textit{meta-features} (division, day, etc.).
    }
    \label{fig:eg}
    \vspace{-0.2in}
\end{figure}

Our goal is to design a novel solution that not only unifies the data modalities under a single, flexible model capable of managing complex relations based on directed multi-graphs, but also offers broad applicability across the domains mentioned earlier and potentially beyond. To this end, we represent the relational information with a node- and edge-attributed directed multi-graph, and the auxiliary or metadata as tabular meta-features. (See Figure~\ref{fig:eg}.) Our proposed solution, \method (for \underline{A}nomaly \underline{D}etection of \underline{A}ttributed \underline{M}ulti-graphs with
\underline{M}etadata), is a unified neural network framework that learns an expressive graph-level representation for directed and attributed multi-graphs and then fuses it with the meta-features within a shared embedding space before feeding the joint embedding to an unsupervised anomaly detection objective. 
Notably, our objective is crafted to handle data heterogeneity; where for example, the journal entries may form multiple clusters  (e.g. purchases vs. interest gains), and individual behaviors can reflect socio-demographic groups (e.g. single vs. married-with-children). Specifically, we replace the classic SVDD objective that aims to learn embeddings tightly centered around a single centroid \cite{tax2004support}, and instead employ an unsupervised loss to accommodate multiple centroids.

The literature is abound with anomaly detection techniques \cite{aggarwal2017introduction,pang2021deep,gupta2013outlier,choi2021deep,akoglu2015graph,ma2021comprehensive}, where a vast body focuses on uni-modal data. Numerous prior work address outliers in tabular data \cite{aggarwal2017introduction,pang2021deep}, possible due to its wide presence in industry and its efficient storage in databases.
However, molding real world anomaly detection problems to tabular outlier detection requires ``flattening'' data from all modalities into manually-extracted features via laborious and often costly domain-expertise \cite{akoglu2021anomaly}. 

On the other hand, graph anomaly detection has been studied mainly on a single graph for detecting node/edge-level anomalies, with much less emphasis on graph-level anomalies \cite{akoglu2015graph,ma2021comprehensive}. 
Few existing traditional approaches to attributed multi-graph anomalies \cite{lee2021gawd,nguyen2023detecting} that are not neural network based are not learnable, restricted to handling single-value edge features, not scalable for larger graphs, and do not take auxiliary metadata into account.
Similarly, the more recent neural network based models \cite{zhao2023using,qiu2022raising,zhao2022graph,zhang2022dual} are not designed to accommodate directed multi-graphs or graphs with metadata as in this work. (See related work in \ref{sec:related} for details.)
Finally, we argue that a straightforward two-stage approach is na\"ive and nontrivial; the reasons are first, treating data modalities/sources separately misses the opportunity to capture inter-dependencies and second, the problem of how to combine multiple anomaly rankings/scores open many possibilities without a principled way to choose in the absence of any labels. 




We summarize our main contributions  as follows.

\cbit

\item {\bf Anomaly Detection in Real-World Settings with Complex Data:~} 
We formulate anomaly detection under data complexity/variety, exhibiting relational as well as auxiliary information, in an elegant framework that can jointly handle complex graphs with node/edge attributes, edge multiplicities, directions and self-loops, meta-features, as well as data heterogeneity. The formulation is driven by anomaly detection problems from two different real-world domains, namely accounting and human mobility, yet is general to apply to possibly other domains.

\item  {\bf A Unified Detection Model:}  We introduce \method, a novel neural network architecture that can digest the aforementioned multi-modal data toward anomaly detection in a unified fashion. It tackles edge multiplicities through set representation learning, employs expressive graph-level embedding that is fused with meta-features in a learned shared embedding space, and finally, optimizes an unsupervised anomaly loss that can accommodate heterogeneous data with multiple latent underlying clusters.



\item  {\bf Generality and Applications:~} \method offers a general framework, where the architecture can be extended to several other domains with data variety, using the idea of learning joint/shared-space embeddings and end-to-end anomaly loss optimization. 
Besides addressing the data variety challenge of big data, our \method also targets business and societal value, as it is applied to two high-stakes domains; accounting (finance) and human mobility (urban). 
Through extensive experiments, 
we show that two-stage solutions are blind-sided and that \method outperforms those as well as other existing baselines significantly on accounting data from three different firms, as well as human GPS trajectory simulations. 
\ceit

{\bf Reproducibility.~} To foster future work on anomaly detection on complex multi-graphs with metadata as well as for practical applications, we open-source the code for \method at \url{https://github.com/konsotirop/ADAMM}.

%% file: 02prelim.tex

We consider anomaly detection on a large database $\mathcal{G} = \{(G_1,M_1) , \ldots ,(G_n,M_n) \}$  of $n$ pairs of directed, node/edge attributed, multi-graphs (multiple edges may exist between two endpoints), and their associated metadata-level features. 
\definition{(Directed, attributed, multi-graph). A graph $G_i = (V_i,E_i,\tau)$ is a directed, attributed, multi-graph, endowed with a function $\tau : V_i \mapsto \mathbb{R}^{d}$ that assigns a real-valued feature vector to every node in $G_i$. Moreover, $E_i$ is a multi-set, where an element $e_t = (u,v,\bold{f}_t)$ is a directed edge between nodes $u$ and $v$ associated with an edge-feature vector $\bold{f}_t \in \mathbb{R}^k$. 
\definition{(Metadata).} Each graph $G_i$ is associated with a vector $\bold{Z_{M_i}} \in \mathbb{R}^{d_M}$ reflecting tabular features.

Usually, we operate on sets (or multi-sets) of variable lengths, where there is no specific order of the elements. In such cases, we need functions that are permutation invariant.
\definition{(Set-function).} A function $f$ acting on sets is called a \textbf{set function} if it is permutation invariant to the order of objects in the set. That is, for any permutation $\pi : f(\{x_1, \ldots ,x_n\}) = f(\{x_{\pi(1)}, \dots , x_{\pi(n)}\}$.

Neural network architectures, like \textsc{DeepSet} \cite{deepset2017}, can implement arbitrary set functions, while the work of Xu et al. \cite{xu2018how} extends such functions for multi-sets.

\textbf{Graph Neural Network (GNN) model:} We use the provably expressive GIN model of \cite{xu2018how}, where the embedding of a node $v$ is updated during the $l^{th}$ layer/iteration using the following aggregation function:
\begin{equation}
\label{eq:gine_conv}
    \bx_{v}^{(l)} = MLP^{(l)}((1+\epsilon) \cdot \bx_v^{(l-1)} + \sum_{u \in \mathcal{N}(v)} \text{ReLU}(\bx_u+\blf_{vu}); \blt_l
    )
\end{equation}
where $MLP$ is a multi-layer perceptron, $\epsilon$ a learnable parameter, $\mathcal{N}(v)$ the neighborhood of node $v$, $\blf_{uv}$ is the feature vector of edge $(u,v)$ and $\blt_l$ a vector of trainable parameters. 

To obtain a graph-level representation $\bZ_{G}$ for the whole graph $G$ we can use a permutation-invariant function \textsc{READOUT} that aggregates node embeddings after the final layer/iteration $L$, i.e.,
\begin{equation}
\label{eq:readout}
\bZ_{G} = READOUT(\{\bx_v^{(L)} | v \in V \}) 
\end{equation}

Our problem can be defined (informally) 
 as follows: 
\begin{tcolorbox}
\begin{problem}(Anomaly Detection of Attributed Multi-graphs with Metadata (ADAMM).) \underline{Given} a database $\mathcal{G} = \{(G_i,M_i)\}_{i=1}^{n}$ of $n$ node- and edge-attributed multi-graphs and their associated metadata; the goal is to \underline{identify} the abnormal (graph, metadata) pairs that differ significantly from the majority in the database.
\end{problem}
\end{tcolorbox}


%% file: 03method.tex
\begin{figure*}[!t]
    \centering
\includegraphics[width=0.9\textwidth,height=3in] {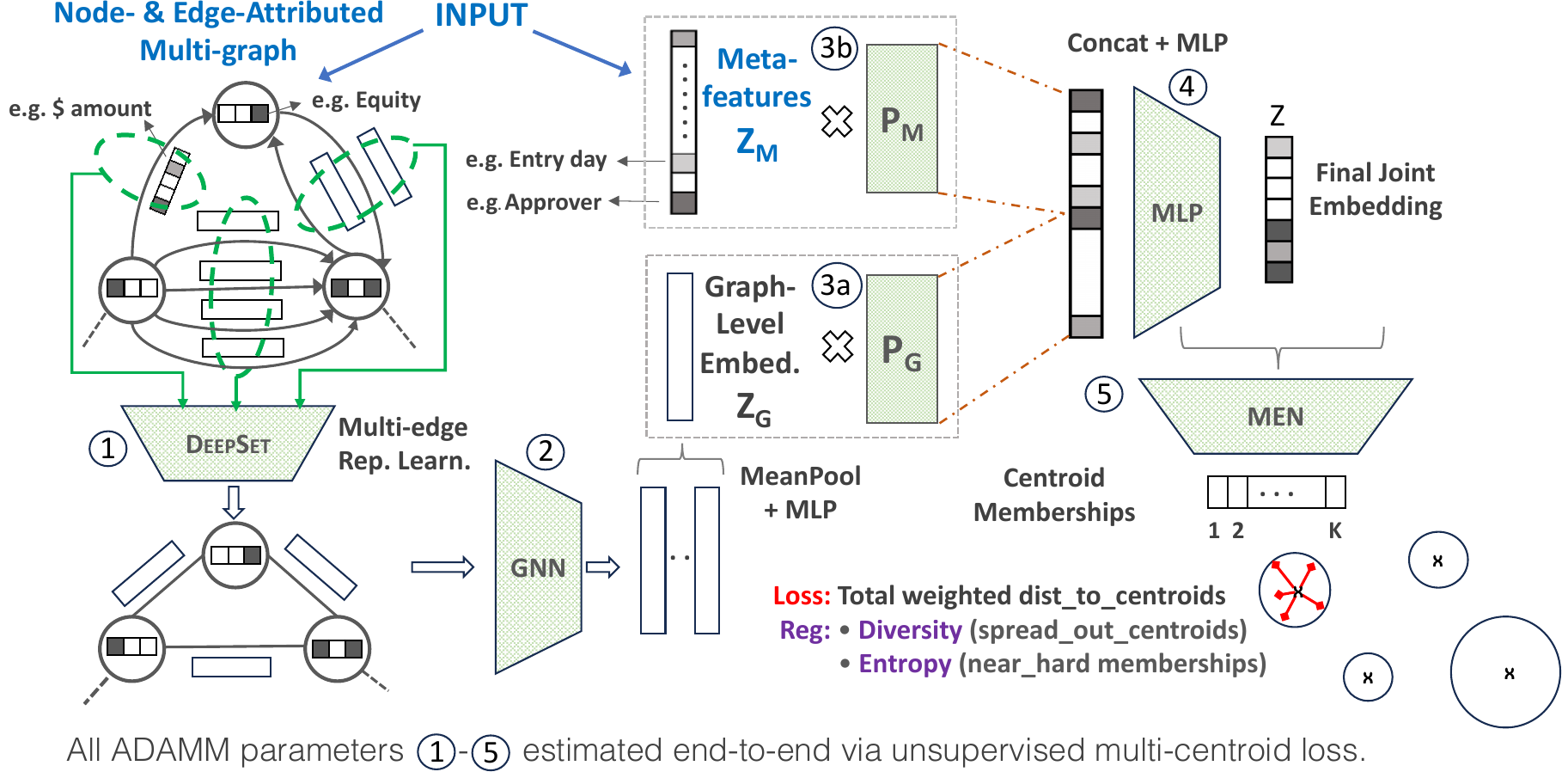} 
    \vspace{-0.1in}
    \caption{A workflow overview of the \method architecture.
    Given two-pronged input (in blue), i.e. attributed multi-graph and metadata, \method first processes the former by \textcircled{1} learning a multi-set representation of the multi-edges, and \textcircled{2} flattening the resulting graph via GNN into node representations that are pooled into a graph-level embedding. Then, \textcircled{3} graph-level embed. and meta-features are projected, \textcircled{4} followed with  a joint embedding learning. Finally, \textcircled{5} the output layer employs an unsupervised regularized multi-centroid  anomaly loss where soft assignments are learned via a membership estimation network (MEN). 
    It is notable that \method provides a unified multi-modal framework
    where parameters of all the modules (in green) are estimated end-to-end.
    }
    \label{fig:overview}
\end{figure*}

\subsection{Data Representation}
\label{ssec:rep}
The input to  \method is a database comprising of pairs of graphs and their associated metadata vectors. In what follows, we describe the capabilities of it in representing complex graph data, the fusion of them with metadata vectors, as well as two concrete examples from the accounting and human mobility domains where this unified representation can be used to model real-world scenarios. 

{\bf Graph representation.~} 
\method is designed to handle complex graph data of virtually any type. Specific design choices allow the model to be able to represent:
\begin{enumerate}
\item \emph{Node attributes (or Node labels):} Nodes can have attributes that are updated after each graph convolution step. When nodes do not have attribute vectors but categorical labels, we can learn representations for these node labels using an embedding layer. 
\item \emph{Edge features:} Edges can have features containing important information about the link between two nodes in the graph. Thus, node representations are updated taking into account not only the embeddings of the neighboring nodes, but also those of incident edges (see also Eq.~\eqref{eq:gine_conv}).
\item \emph{Edge direction:} In various domains as with transactions (accounting) and trips (mobility), edge direction is semantically important. Thus, we enhance edge features with an encoding of the direction of the edge. Specifically, for each multi-edge $(u,v,\bold{f}_t)$, between nodes $u$ and $v$, we encode it using label $``1"$ if the edge is present in the graph, i.e. $(u,v,\bold{f}_t, ``1")$, and in the meanwhile, augment another reversed edge $(v,u,\bold{f}_t, ``2")$ with label $``2"$ into the graph. Also, we reserve  label $``0"$ for self-loop edges, i.e. $(u,u,\bold{f}_t)$ becomes $(u,u,\bold{f}_t, ``0")$. These edge direction labels are then applied as input to an embedding layer that produces the edge direction representation vector $\bold{d}_t$. The final representation vector $\bold{f}_t'$ for the edge $e_t$ is then obtained as the sum of the edge features vector $\bold{f}_t$ and of the edge direction vector, that is, $\bold{f}_t' = \bold{f}_t + \bold{d}_t$.
\item \emph{Multiple edges:} Currently, GNNs are not able to handle multi-edges. Instead, they assume there exists a unique edge between two nodes, as in Eq.~\eqref{eq:gine_conv} for GIN convolution. Multi-edges, however, model the multiple interactions that can occur between two nodes in a network and each has its own feature vector.  \method is designed to handle multi-edges by learning a single edge representation from the multi-set of edges. More precisely, we treat the edge features of the multi-edges $F = \{\blf_1', \ldots ,\blf_T'\}$ as a multi-set and we use a permutation-invariant multi-set function $f : F \mapsto \mathbb{R}^{d_e}$ to learn an edge-level $d_e$-dimensional representation vector.
\end{enumerate}

The versatility of \method in handling complex graph-data allows it to be used in a wide variety of domains. We present two exemplar ones as follows.
\begin{itemize}
\item[(i)] Bookkeeping Graphs \cite{liang2023bookkeeping}: Each graph is a representation of a \textit{journal entry}: a detailed transaction record. Every account present in the entry is associated with a node, with its label being the account type (e.g. equity, revenue, etc.). A directed edge represents monetary flow from a credited account to a debited account and the feature of an edge is the monetary value associated with this transaction. Directed multi-edges capture multiple credit/debit flows that can take place between two accounts. 
\item[(ii)] Human Mobility (or Activity) Graphs \cite{schneider2013unravelling}: These represent the mobility or activity behavior of an agent within a time-frame (e.g., a day of the week). Nodes represent visited locations, while node labels represent the Points Of Interest (POI) type in that location (school, restaurant, etc.). Directed multi-edges stand for the trips between those locations and their features capture information about the trip (duration, distance, etc.).
\end{itemize}

{\bf Metadata representation.~} 
\method is able to fuse the graph-level representation  with associated metadata vectors that contain auxiliary information. We give examples of such information in the two aforementioned domains below.

\begin{itemize}
\item[(i)] Metadata for Bookkeeping Graphs: The metadata vector contains information regarding the ID of the user that created the specific journal entry, the approver of this entry, total credit amount, a binary indicator of whether it is a reversal, the date transactions took effect, or the date transactions were recorded in the journal, etc.
\item[(ii)] Metadata for Human Mobility Graphs: For activity graphs the metadata vector could contain information about the day of the week this activity took place (e.g., Tuesday), a vector representation of the agent it describes (or simply a unique ID), or other information that could contain GPS related information, like speed-limit violations, etc.  
\end{itemize}

\subsection{A Unified Neural Network Architecture}
\label{ssec:arch}

\method provides a unified architecture for anomaly detection in a database of graphs and their associated metadata features. Figure~\ref{fig:overview} presents an overview of our model and the steps it involves. The input is two-pronged: a directed, node/edge attributed multi-graph and its associated metadata vector. Following the set representation learning of the multi-edges, a GNN is employed to learn a graph-level embedding, which is then fused with the metadata vector to obtain the final joint embedding. A parameter estimation network decides on the (soft) membership of the final embeddings to one of $K$ clusters. \method is trained in an end-to-end fashion, i.e. all of its parameters are optimized jointly with respect to a suitable objective function that minimizes the weighted distance of embeddings to the $K$ centroids of the clusters. Additional regularization terms are introduced to spread-out the centroids as well as to nudge the estimation network toward more confident assignments of cluster memberships.
We describe each of these steps in greater detail next.

\subsubsection{Graph-level Embedding}
\label{sssec:gemb}
We learn a graph-level embedding in two steps:
\cbit

    \item Multi-edge Representation Learning: As noted, GNNs (including GIN) can \textbf{not} readily handle multi-edges. For this reason, we ``flatten" all directed multi-edges between two nodes to a single undirected edge and its associated feature vector by learning a permutation-invariant multi-set function based on a DeepSet\cite{deepset2017} architecture. This procedure is depicted in Figure~\ref{fig:overview} (see step \textcircled{\raisebox{-0.9pt}{1}}), where the input attributed  multi-graph is transformed using a learnable multi-set function to an attributed graph with single edges among its pairs of nodes.
    \item Node Embeddings: The transformed graph, where all multi-edges have been replaced by an attributed single edge, is used as input to a GNN model (step \textcircled{\raisebox{-0.9pt}{2}} in Figure~\ref{fig:overview}). In \method we opt to use GIN \cite{xu2018how} as a  provably expressive GNN. GIN learns node embeddings by performing the graph convolution of Eq.~\eqref{eq:gine_conv}, also incorporating the  edge features.
    \ceit
     To obtain a graph-level embedding we use a \textsc{READOUT} function on the node embeddings (see Eq.~\ref{eq:readout}). We implement this function by performing mean pooling over the node embeddings followed by a multi-layer perceptron (MLP) to learn the final graph-level embedding as
    \begin{equation}
    \label{eq:graph_emb}
    \bZ_{G} = MLP\left(\frac{1}{|V|}\sum_{ v \in V} \bx_v^{(L)} ; \blt_G \right) \;.
    \end{equation}

\subsubsection{A Unifying Embedding Space for Graph and Metadata}
\label{sssec:uni}
After having obtained a graph-level embedding, we learn a joint representation of the graph and its metadata in a unifying embedding space. We are motivated by the CLIP-style latents \cite{ramesh2022hierarchical} that learn a shared embedding space for images and their associated text captions, analogous to our graph and metadata pairs. More precisely, we first linearly project  the graph-level embedding vector $\bZ_G$  as well as the metadata vector $\bZ_M$ by learning two projection functions (with separate parameters) 
$P_G(\bZ_G ; \blt_G): \mathbb{R}^{d_G} \mapsto \mathbb{R}^{d_P}$ and $P_M(\bZ_{M} ; \blt_M) : \mathbb{R}^{d_M} \mapsto \mathbb{R}^{d_P}$
to obtain two new vectors $\bZ_G'$ and $\bZ_M'$ of the same length $d_P$ as they share the same space. We normalize both vectors to have unit $l_2$ norm and then concatenate them. Finally, we employ an MLP to obtain the  final joint embedding, denoted $\bZ \in \mathbb{R}^{d}$ (see steps \textcircled{\raisebox{-0.9pt}{3}} - \textcircled{\raisebox{-0.9pt}{4}} in Figure~\ref{fig:overview}) as 
\begin{equation}
\label{eq:final}
\bZ = MLP\left(\text{CONCAT}(\bZ_G',\bZ_M') ; \blt_J \right) \;.
\end{equation}


\subsection{Anomaly Detection Loss}
\label{ssec:loss}

Objective functions used in anomaly detection, like One-Class DeepSVDD \cite{ruff2018deep}, make the somewhat strong assumption that all of the normal instances come from the same distribution. Hence, their objective is to use a deep neural network to map all normal instances as close to the center of a \textit{single} hypersphere, with anomalous instances identified as those mapped farther from this centroid. However, this objective does not take into account the multiple modalities or heterogeneities that may exist in real world data. For this reason, we introduce a new objective function that accommodates \textit{multiple} clusters of the input samples. \method estimates the (soft) cluster membership of each sample using a membership estimation network and tries to minimize the total average weighted distance from the $K$ centroids, where hyperparameter $K$ is carefully tuned (as discussed later in \ref{ssec:selection}). Contrary to One-Class DeepSVDD, where the center of the hypersphere is fixed during the training process,  the $K$ centroids in our case are inferred from the membership estimation network and the final embedding vectors. 

{\bf Membership Estimation Network (MEN).~}
The MEN is an MLP with a softmax activation function (step \textcircled{\raisebox{-0.9pt}{5}} of Fig.~\ref{fig:overview}) that gives the membership predictions for the final embedding vectors $\bZ$ in Eq. \eqref{eq:final}.  That is, 
\begin{equation}
\label{eq:gamma_eq}
    \widehat{\blg} = \text{softmax}(\text{MLP}(\bZ ;\blt_{MEN})) \;,
\end{equation}
where $\widehat{\blg}$ is a $K$-dimensional vector depicting the soft membership probability predictions of $\bZ$. 

In what follows, and for a batch  of $N$ pairs of (graph, metadata) samples, where $\bZ_i$ is the embedding of the $i^{th}$ sample, we denote by $\widehat{\blG}$ the $N\times K$ matrix of cluster membership estimations from Eq.~\eqref{eq:gamma_eq} and by $\widehat{\gamma}_{ik}$ each entry of this matrix. 

Then, the cluster centroids $\widehat{\blc}_k \in \mathbb{R}^{d}$ can be calculated  using the cluster membership estimations from Eq.~\eqref{eq:gamma_eq} and embedding vectors $\bZ_i$, for $i=1, \ldots, N$, by
\begin{equation}
\label{eq:centroids}
\widehat{\blc}_k = \frac{\sum_{i=1}^{N} \widehat{\gamma}_{ik} \bZ_i}{\sum_{i=1}^{N} \widehat{\gamma}_{ik}} \;.
\end{equation}



{\bf Loss Function and Anomaly Score.~}
Having estimated the embedding vectors and cluster membership estimations, the anomaly score of a sample $T_i=(G_i,M_i)$ is then defined as the weighted sum of the Euclidean distance between the final embedding vector $\bZ_i$ and the cluster centroids $\widehat{\blc}_k$'s, i.e. 
\begin{equation}
\label{eq:anomaly_score}
\text{score}(T_i) = \sum_{k=1}^{K} \; \widehat{\gamma}_{ik} \lVert \bZ_i -\widehat{\blc}_k \rVert^2 \;,
\end{equation}
which also serves as the \emph{anomaly score} of a sample $T_i$ (the higher, the farther and the more anomalous).

\method is then trained in an end-to-end fashion to optimize the following unsupervised objective.
\begin{equation}
\label{eq:objective}
    \min_{\blt} \frac{1}{N}\sum_{i=1}^{N} \sum_{k=1}^{K} \widehat{\gamma}_{ik} \lVert \bZ_i -\widehat{\blc}_k \rVert^2 + \lambda_1 \cdot H(\widehat{\blG}) + \lambda_2 \cdot D(\widehat{\blC})
\end{equation}
where $\bm{\theta}$  depicts  all \method parameters collectively and $\lambda_1, \lambda_2$ are hyperparameters that aim to strike a balance between the distance-to-centroids anomaly loss (first term) and two regularization terms, respectively, Entropy and Diversity, which we describe as follows.
\cbit
\item The first is an entropy regularization that forces the network to be more confident on the cluster to which it estimates an input sample to belong. More specifically, we aim to minimize the average entropy over the rows of the $\widehat{\blG}$ membership estimation matrix as
\begin{equation}
H(\widehat{\blG}) = \frac{1}{N}\sum_{i=1}^{N}\sum_{i=1}^{K} -\widehat{\gamma}_{ik}\log(\widehat{\gamma}_{ik}) \;.
\end{equation}
\item The second is a diversity  term that promotes separation between the cluster centroids to avoid the undesired mode-collapse solutions where the network collapses all centroids to the same point. Letting $\widehat{\blC}$ depict the $K \times d$ matrix containing the cluster centroids $\widehat{\blc}_k$'s as its rows; 
\begin{equation} 
D(\widehat{\blC}) = -\log(\det(\text{Cov}(\widehat{\blC})) \;, 
\end{equation}
where $\det(\text{Cov}(\widehat{\blC}))$ is the determinant of the covariance matrix of $\widehat{\blC}$. In effect, the larger the determinant of the covariance matrix, the more the centroids are dispersed, promoting separation between and diversity among the cluster centroids. A similar term has also been used in \cite{das2012selecting} for selecting diverse features in regression settings.
\ceit

\subsection{Model Selection}
\label{ssec:selection}

\method, as with other deep neural networks based models, is configured with a set of hyperparameters (HPs), 
such as the number of layers, weight decay and learning rates, number of training epochs, among others.
In addition, our multi-centroid anomaly objective incurs 
the number of centroids $K$, and the $\lambda_1$ and $\lambda_2$ terms from Eq.~\eqref{eq:objective}.  Each different configuration of these  results in a different model, with potentially drastic differences  in anomaly detection effectiveness. The challenge is that anomaly detection is an unsupervised task, where we usually lack ground-truth labels of whether a sample is an anomaly. As a result, we do not have a labeled validation set for hyperparameter tuning. For this reason, we devise an unsupervised validation score, \emph{without using any labels}, toward selecting an effective model that performs better in anomaly detection than what we would have obtained by picking at random (in absence of any other guidance). 

Given a family of models, $\mathcal{M}$, we opt to choose the model $m$ that minimizes the sum of the weighted distances of the $N$ samples in the training set from the $K$ centroids  as our model selection criterion, specifically, 
\begin{equation}
\label{eq:model_selection}
\sum_{i=1}^{N} \sum_{k=1}^{K} \; \widehat{\gamma}_{ik} \lVert \bZ_i -\widehat{\blc}_k \rVert^2 \;.
\end{equation}
This rule favors the model that succeeds into learning a tight representation of the training instances into each of the $K$ clusters by better extracting their shared patterns, which in effect helps reveal the anomalies that deviate from these patterns.
In experiments, we compare the effectiveness of our model selection criterion against random picking (i.e. average/expected performance over possible HP configurations).


%% file: 04experiments.tex
\subsection{Experimental Setup}
\label{ssec:setup}

{\bf Datasets.~} 
For evaluation we use four datasets from two different domains, each containing a large database of graphs and their associated metadata. 
Those include annual general-ledger journal entries from three different firms, in collaboration with PwC. The fourth dataset involves simulated human GPS trajectories.
Summary of datasets is given in Table \ref{tab:datasets}. 
\begin{itemize} 
\item \emph{Accounting Datasets}: Three datasets from accounting consist of all annual journal entries from different firms anonymized as SH, HW, and KD. Each dataset contains tens of thousands of bookkeeping graphs \cite{liang2023bookkeeping} capturing itemized transactions between impacted accounts along with dollar amounts, and metadata entries capturing auxiliary journal information including entry and effective date, requester, approver,  reversal indicator, and so on.
\item \emph{Human Mobility Dataset}: The fourth dataset, referred to as MobiNet, contains the trajectories of $10,000$ simulated agents over a period of two weeks. We use these trajectories to extract daily activity graphs \cite{schneider2013unravelling} of the places (i.e. POI) visited by each agent and the trips between them, as described in Section~\ref{ssec:rep}. The metadata contains information about individual trips and are transformed to a single vector using a \textsc{DeepSet} architecture during the end-to-end training of \method.
\end{itemize}

\begin{table}
\centering
\caption{Dataset Summary Statistics}
\vspace{-0.05in}
\label{tab:datasets}
\resizebox{0.98\linewidth}{!}{\begin{tabular}{p{0.9cm}|r|r|r|r|r|r}
\toprule
Name  & Graphs & Nodes & Multi-edges & Node-attr. & Edge-attr. & Meta-feat.\\ 
\midrule
SH & 39,011 & [1,15] & [1,338] & 11 & 1 & 11 \\
KD & 152,105 & [1,91] & [1,774] & 10 & 1 & 9 \\
HW & 90,274 & [1,25] & [1,897] & 11 & 1 & 7 \\
MobiNet & 140,000 & [1,22] & [1,59] & 41 & 4 & 9 \\ 
\bottomrule
\end{tabular}}
\vspace{-0.15in}
\end{table}

{\bf Baselines.~} For comparison, we use as baselines existing graph-level anomaly detectors and tabular data outlier detectors. Unlike \method, existing graph-level anomaly detectors can not handle multi-edges. Therefore, we collapse all multi-edges to a single edge and use the average representation of their feature vectors. To the best of our knowledge, there is also no prior work that fuses graphs and metadata and provides a single anomaly score. For this reason, we employ two-stage baselines: First, we create a ranking of the samples with respect to their anomaly score as obtained by a graph-level anomaly detector. Then, we create a second ranking by using a tabular data outlier detector. We combine these two rankings to obtain a single graph\&metadata anomaly ranking using two well-established aggregation methods detailed as follows.
\begin{itemize}
\item[($a$)] \textit{Graph-level Anomaly Detectors}: We first aim to detect graph-level anomalies using the following baselines: 
\begin{itemize}
\item[(1)] Weisfeiler-Lehman (WL) graph kernel \cite{shervashidze2011weisfeiler}, followed by the OCSVM outlier detector \cite{taylor2000support} that can admit a kernel matrix as input.
\item[(2)] graph2vec \cite{narayanan2017graph2vec}, for graph-level embedding, followed by the OCSVM detector. 
\item[(3)] DOMINANT \cite{ding2019deep}, a GNN-based node anomaly detector, from which we average the scores to obtain a graph-level anomaly score.
\end{itemize}
\item[($b$)] \textit{Tabular Data Outlier Detectors}: 
We use the tree-ensemble based Isolation Forest  algorithm \cite{liu2008isolation} to score outlierness on the  meta-features, which is the state-of-the-art tabular data outlier detector \cite{emmott2015meta}. 

\end{itemize}

After having obtained a ranking from ($a$) the graph-level anomaly detectors and ($b$) the tabular data anomaly detectors, we create a unique ranking for pairs of graphs and metadata, by using:
($i$) a BFS-style aggregation that first sorts the results of each stage in descending order of their anomaly score, and then selects the next object that has the highest anomaly score by visiting the lists in a BFS fashion \cite{lazarevic2005feature}; and 
($ii$) the Inverse Rank (IR) aggregation method, in which we score each sample by $\frac{1}{r_a} + \frac{1}{r_b}$, where $r_a$ is the rank by the graph-level anomaly detector and $r_b$ by the tabular data outlier detector.  

Overall we construct 6 baselines based on (1)--(3) $\times$ ($i$)--($ii$).

{\bf Labeled Anomalies.~}
Our datasets do not come with any ground truth anomalies, therefore, we use the guidance of experts in the fields of accounting and human mobility to simulate anomalies that are typically present in these datasets and of interest in detecting them. We create two types of graph-level anomalies, as well two types of metadata-level anomalies. 

\input{TABLES/auroc}

\textbf{1) Graph anomalies} involve small perturbations in nodes and edges as follows.
\cbit
\item \textit{Label change} (\textbf{GA1}): We change the label of a node to a randomly chosen new label. This injection corresponds to entry-error in an accounting dataset, or an unusual visit to a new POI in human mobility behavior.
\item \textit{Path injection} (\textbf{GA2}): We delete an edge between nodes $u$ and $v$ and rewire through an intermediary, creating a path $u$-$z$-$v$. This injection  mimics money-laundering in finance, where funds are passed through an intermediate account instead of being transferred directly. For human mobility, this corresponds to an unusual stop.
\ceit
\textbf{2) Metadata anomalies} perturb feature values and reflect different semantics in both domains.  \\
\indent For Accounting Datasets:
\cbit
\item \textit{Unusual back-dating} (\textbf{MA1}): We pick a subset of entries with effective date close to the entry date (up to 3 days before), and change the entry date randomly to one of $\{7,14,21\}$ days after the effective date. This corresponds to an unusual late entry date. We lack information about entry date in HW, hence our experiments do not use this type of anomaly for this dataset.
\item \textit{Combination of unrelated transactions} (\textbf{MA2}): We merge two unrelated transactions by creating a new one with a unique Journal ID. We set the metadata entries to a randomly chosen value from the two initial transactions. Note that this injection also modifies the graph structure into the representation of the merged journals.
\ceit
\indent For Human Mobility Dataset:
\cbit
\item \textit{Unusual start time} (\textbf{MA3}): We change the start time of a trip to a very early (or late one). This corresponds to a trip occurring in an unusual time.
\item \textit{Unusual trip duration} (\textbf{MA4}): We change the duration of a trip to an unusually long one. 
\ceit

\textbf{3) Potpourri anomalies}  involve a combination of graph and metadata level anomaly injections, where we pick one graph level anomaly and one metadata level anomaly from above and inject \underline{both} to a sample. 

We inject anomalies on $5\%$ of the samples in each dataset, where we use half of the original dataset for training, and the remaining half with the injected anomalies for testing.
Note that the labeled anomalies are used only for evaluation purposes and not during model training or model selection.

\textbf{Model Selection.~} Anomaly detection is typically a fully unsupervised task, where we lack ground truth labels of which samples are anomalous. As a result, we do not have a validation set for hyperparameter tuning. For this reason, for each of the baseline methods, we consider a set of hyperparamater configurations, across which we report the average performance.  This corresponds to the expected performance of each method if one were to select a configuration at random. For \method we show that our proposed model selection criterion presented in \ref{ssec:selection} can consistently yield better results than what we would expect when picking hyperparameters at random (in the absence of any other guidance). 

\textbf{Hyperparameter (HP) Configurations:} Detailed HP configurations can be found in the appendix \ref{sec:appendix}.

\subsection{Detection Results}
\label{ssec:results}
In evaluating proposed \method, 
we conducted a series of experiments to answer the following questions:
\begin{itemize}
\item[\textbf{Q1) Effectiveness:}] How effective is \method in detecting graph- and metadata-level anomalies, as  compared to the two-stage baseline approaches?
\item[\textbf{Q2) Model Selection:}] 
Can our proposed unsupervised model selection criterion for \method select a model (i.e. hyperparameter configuration) that is better than random picking (i.e. avg. performance across config.s)?
\item[\textbf{Q3) 
Ablation: }] How important are key components of \method in the detection results?
\end{itemize}

\textbf{A1:} To answer the first question, we conduct extensive experiments using all four datasets and graph, metadata as well as potpourri anomalies. The results are presented for each method across all datasets and injection types in Table~\ref{tab:full_results}  based on the Area Under the Receiver Operator Characteristic Curve (AUROC), and in Table~\ref{tab:ap_results} based on the Area Under Precision-Recall Curve (AUPRC). We observe that \method succeeds in detecting both graph-level and metadata-level anomalies effectively. It outperforms the baseline methods in $3$ out of $4$ datasets and across all injection types, where the baselines do not show consistent performance. \method performs consistently well for all types of anomalies, whether they are graph, metadata-level, or even of mixed type. The only exception is the MobiNet dataset, where the nature of metadata information (multiple vectors for a single graph that have to be aggregated) poses significant challenges\footnote{For baseline methods, we score all vectors separately and assign the maximum score as the anomaly score of the sample. This gives better results than taking the average as the latter dilutes the signal among multiple vectors.}

The superior performance of \method over baselines is also validated using the Wilcoxon signed rank test. \method not only has the lowest average rank among the competitors, but is also significantly better at p-value $p=0.05$.

\input{TABLES/auprc}

\begin{figure}[!th]
\vspace{-0.15in}
    \centering
\includegraphics[width=0.85\linewidth] {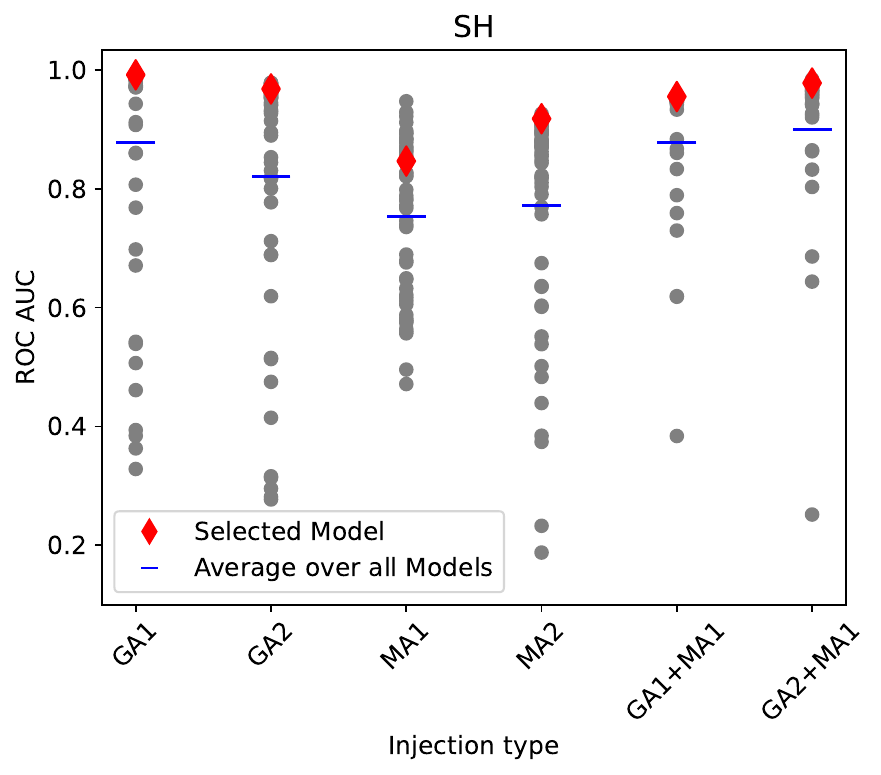}
\vspace{-0.1in}
    \caption{Model selection for \method over all models with different hyper-parameter configurations. The model selected consistently performs better than random picking, i.e. average/expected performance over all models. 
    }
    \label{fig:model_selection}
\end{figure}

{\bf A2:~} 
The problem of model selection is an important one in unsupervised anomaly detection. The lack of labels and of a validation set makes it challenging to choose an effective model. For \method we provide a validation criterion tightly connected to its loss function (recall Eq.~\eqref{eq:model_selection}). In Figure~\ref{fig:model_selection} we see that the model \method chooses based on its unsupervised criterion achieves  consistently better performance than random picking that corresponds to the average performance across all configurations. This makes \method not only able to spot graph and metadata level anomalies, but also robust against different hyperparameter choices.

{\bf A3:~} 
\method exhibits three key building blocks; multi-edge representation learning, graph-metadata fusion, and a suitable anomaly detection loss. Accordingly, we perform an \emph{ablation study} and design threevariants of \method, each excluding the respective design component to demonstrate its added benefit.

\input{TABLES/ablation}

\begin{itemize}

\item[V1.] \textbf{\method without Metadata Fusion}:
Here we remove the metadata fusion component and instead we input only the graph-level embeddings $\bZ_G$ to the membership estimation network. Our goal is to explore if the metadata component interferes with the graph-level component by having a negative influence on graph-level anomaly detection when only such anomalies are present.

\item[V2.] \textbf{\method without DeepSet}: In this version, we remove the DeepSet component that aims to learn a single representation of the attributed multi-edges. Instead, we simply average the attributes over each multi-edge.

\item[V3.] \textbf{\method with One-Class DeepSVDD loss}: Finally, we compare $\method$ and its loss function in Eq.~\eqref{eq:objective} with a varint where we replace it with the loss introduced by the One-Class DeepSVDD\cite{ruff2018deep} method which, as we described in \ref{ssec:loss}, maps all normal instances to a \emph{single} hypersphere centered around a \emph{fixed} centroid.
\end{itemize}

Results of the ablation study are given in Table ~\ref{tab:ablation} for SH (as a representative of the transaction datasets) as well as the MobiNet dataset.
We see that \method outperforms all of its variants on the $SH$ dataset, demonstrating the importance of the various components in anomaly detection. The improvement is particularly noticeable for the MA2 type anomalies (merge of unrelated transactions), which is of mixed type (both metadata and graph).  We note that \method without metadata also performs well for graph-only anomalies of type GA1 and GA2. In fact, excluding metadata lifts the interference on MobiNet, leading to better detection.

\subsection{Case Studies}
\label{ssec:case}

\begin{figure}[!t]
    \centering
    \begin{tabularx}{\linewidth}{XX}
\includegraphics[width=1.2\linewidth] {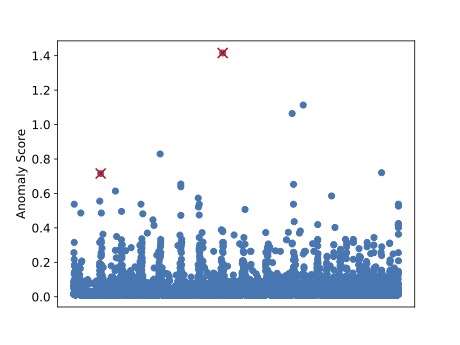} &
    \includegraphics[width=1.1\linewidth] {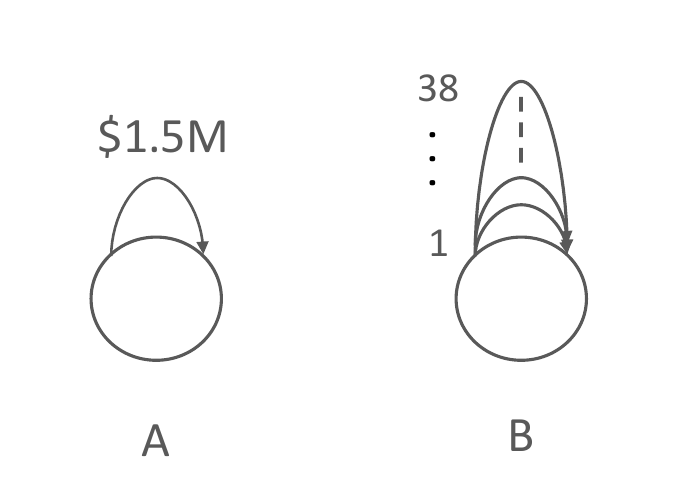}
    \end{tabularx}
    \vspace{-0.15in}
    \caption{
        Analyzing detected accounting anomalies. (Left) Anomaly scores (vs journal ID) of all $39,011$ entries in the SH dataset. (Right) Two example  graphs, A and B, that are identified as anomalous by \method in SH.  
    }
   \vspace{-0.2in}
    \label{fig:transaction_examples}
\end{figure}

Through quantitative experiments in \ref{ssec:results} we showed that \method can successfully spot expert-guided injected anomalies. To further validate the effectiveness of our method, we  consider the {original} $SH$ dataset that contains \textbf{no} injected anomalies. That is, we use the whole dataset of $39,011$ graphs with metadata for training and inspect their anomaly scores obtained by Eq.~\eqref{eq:anomaly_score}. As presented in Figure~\ref{fig:transaction_examples} (left), we see that \method is able to highlight a small fraction of the samples as standing out from the majority. 

As \method is unique in handling complex directed graphs with attributes and multi-edges, we take a closer look at two example graphs as shown in  Figure~\ref{fig:transaction_examples} (right).
Self-loops are the common feature of these graphs: the first has one self-loop with a large dollar amount (\$1.5M), while the second contains 38 self-loops in one graph. From an accounting domain perspective, self-loops represent transactions recorded by moving dollars \emph{within} the same general ledger (GL) account. From a bookkeeping standpoint, these within-GL movements indicate the presence of misidentification of the correct sub-ledger account in the recording of  a prior transaction. The self-loop in the current journal entry is then designed to correct such a misidentification at a later date. In the first graph A, a total of $\$1.5$M was recorded in a sub-ledger incorrectly, necessitating the current self-loop transaction to correct the cumulative mistakes made earlier. The second graph B (with 38 self-loops) is even more pronounced in terms of the number of corrections involved as well as the presence of errors beyond the simple one illustrated in the first example. 

Based on this transaction-level bookkeeping analysis, these two spotted transactions are indeed unusual and worthy of the auditor's attention to examine further. \method's ability  to spot these anomalies involving edge-attributes and multi-edges can be of assistance to the accounting/auditing practitioners.

%% file: TABLES/auroc.tex
\begin{table*}[!t]
\centering
    \caption{Anomaly Detection Results for all methods across all datasets based on \textsc{AUROC}. For baseline methods we run the experiments over a grid of hyperparameters and report the average performance, along with the std. dev. \method employs a model selection criterion and outputs a unique ranking. Last row reports significance test results, where (**) and (***) denote that \method is significantly better than baselines w.r.t. the Wilcoxon Signed Rank Test at $p=0.05$ and $p=0.01$, respectively. }
    \vspace{-0.05in}
    \label{tab:full_results}
    \setlength{\tabcolsep}{3pt}
    \begin{tabular}{c|c||c||c|c|c|c|c|c}
        \toprule
{Dataset}&
{Anomaly Type}
                    &  \method
                    & WL+BFS
                    &  WL+IR 
                    & G2V+BFS
                    & G2V+IR
                    & DOM.+BFS
                    & DOM.+IR \\
        \midrule
\multirow{6}{*}{SH}
& GA1  & \textbf{0.992}  & 0.925 $\pm$ 0.01 & 0.922 $\pm$ 0.01 & 0.839 $\pm$ 0.09 & 0.833 $\pm$ 0.09 &  0.824 $\pm$ 0.01 & 0.821 $\pm$ 0.01\\
& GA2  &  \textbf{0.968} & 0.827 $\pm$ 0.02 & 0.829 $\pm$ 0.02 & 0.854 $\pm$ 0.02 & 0.854 $\pm$ 0.02 & 0.834 $\pm$ 0.01 & 0.837 $\pm$ 0.01 \\
& MA1  &  \textbf{0.846} & 0.591 $\pm$ 0.02 & 0.610 $\pm$ 0.03 & 0.586 $\pm$ 0.01 & 0.592 $\pm$ 0.01 & 0.602 $\pm$  0.02 & 0.613 $\pm$ 0.02 \\
& MA2  &   \textbf{0.918} & 0.638 $\pm$ 0.02 & 0.642$ \pm$ 0.02 & 0.614 $\pm$ 0.01 & 0.618 $\pm$ 0.01 &  0.615 $\pm$ 0.01 & 0.618 $\pm$ 0.01 \\
& GA1 + MA1  & \textbf{0.955} & 0.899 $\pm$ 0.01 & 0.897 $\pm$ 0.02 & 0.807 $\pm$ 0.01 & 0.800$ \pm$ 0.01 & 0.811 $\pm$ 0.01 & 0.807 $\pm$ 0.02\\
& GA2 + MA1  &  \textbf{0.977} & 0.841 $\pm$ 0.03 & 0.840 $\pm$ 0.01 & 0.871 $\pm$ 0.01 & 0.869$\pm$0.02  & 0.836 $\pm$ 0.02 & 0.838 $\pm$ 0.01\\
        \hline
\multirow{6}{*}{KD}
& GA1  & \textbf{0.928} & 0.885 $\pm$ 0.01 & 0.880 $\pm$ 0.01 & 0.835 $\pm$ 0.04 & 0.828 $\pm$ 0.04 &  0.462 $\pm$ 0.01 &  0.450 $\pm$ 0.01 \\
& GA2  & \textbf{0.939} & 0.825 $\pm$ 0.03 & 0.828 $\pm$ 0.04 & 0.820 $\pm$ 0.01 & 0.824 $\pm$ 0.01 &  0.543 $\pm$ 0.01 &  0.528 $\pm$ 0.01\\
& MA1  &  \textbf{0.841} & 0.727 $\pm$ 0.01 & 0.716 $\pm$ 0.03 & 0.729 $\pm$ 0.01 & 0.718 $\pm$ 0.01 &  0.610 $\pm$ 0.01 & 0.588 $\pm$ 0.01\\
& MA2  & \textbf{0.854} & 0.738 $\pm$ 0.02 & 0.743 $\pm$ 0.02 & 0.736 $\pm$ 0.02 & 0.741 $\pm$ 0.02 & 0.518  $\pm$ 0.01 & 0.505 $\pm$ 0.01 \\
& GA1 + MA1  & \textbf{0.933} & 0.901 $\pm$ 0.01 & 0.895 $\pm$ 0.01 & 0.814 $\pm$ 0.07 & 0.805 $\pm$ 0.01 &  0.458 $\pm$ 0.01 & 0.448 $\pm$ 0.01 \\
& GA2 + MA1  & \textbf{0.916} & 0.849 $\pm$ 0.03& 0.849 $\pm$0.04 & 0.818 $\pm$ 0.01 & 0.805 $\pm$ 0.07 &  0.537 $\pm$ 0.01 &  0.522 $\pm$ 0.01 \\
    \hline
\multirow{3}{*}{HW}
& GA1  & \textbf{0.973} & 0.922 $\pm$ 0.01 & 0.916 $\pm$ 0.01 & 0.922 $\pm$0.03  & 0.920 $\pm$ 0.03 & 0.713 $\pm$ 0.21 & 0.710 $\pm$ 0.21 \\
& GA2  & \textbf{0.994} & 0.895 $\pm$ 0.01 & 0.888 $\pm$ 0.01 & 0.666 $\pm$ 0.05 & 0.660 $\pm$ 0.05 & 0.400 $\pm$ 0.07 & 0.406 $\pm$ 0.07 \\
& MA2  & \textbf{0.967} & 0.691 $\pm$ 0.02 & 0.661 $\pm$ 0.02 & 0.706 $\pm$ 0.02 & 0.694 $\pm$ 0.01 & 0.535 $\pm$ 0.01 & 0.527 $\pm$ 0.01\\
        \hline
\multirow{8}{*}{MobiNet}
& GA1  & 0.526 & 0.676 $\pm$ 0.01 & \textbf{0.678} $\pm$ 0.02 & 0.466 $\pm$ 0.07 & 0.467 $\pm$ 0.05 & 0.320 $\pm$ 0.01 & 0.323 $\pm$ 0.01\\
& GA2  & 0.491 & 0.487 $\pm$ 0.02 & 0.482 $\pm$ 0.03 & 0.499 $\pm$ 0.05 & \textbf{0.505} $\pm$ 0.04 &  $ 0.341 \pm$ 0.01 & 0.346 $\pm$ 0.01\\
& MA3  & 0.441 & 0.558 $\pm$ 0.01 & 0.563 $\pm$ 0.01 & 0.556 $\pm$ 0.02& \textbf{0.574} $\pm$ 0.03 &  $ 0.411 \pm$ 0.01& 0.415 $\pm$ 0.01\\
& MA4  & 0.450 & 0.492 $\pm$ 0.01 & 0.490 $\pm$ 0.01 & 0.517 $\pm$ 0.02 & \textbf{0.524} $\pm$ 0.01 &  $0.349 \pm$ 0.01 & 0.353 $\pm$ 0.0\\
& GA1 + MA3  & 0.563 & 0.750 $\pm$ 0.01 & \textbf{0.754} $\pm$ 0.02 & 0.451 $\pm$ 0.01 & 0.454 $\pm$ 0.02 &  0.326 $\pm$ 0.01 & 0.329 $\pm$ 0.03\\
& GA1  + MA4 & 0.678 & 0.743 $\pm$ 0.02 & \textbf{0.747} $\pm$ 0.01 & 0.457 $\pm$ 0.02 & 0.460 $\pm$ 0.01&  $0.350 \pm$ 0.01 & 0.353 $\pm$ 0.02\\
& GA2 + MA3  & 0.470 & 0.483 $\pm$ 0.01 & 0.480 $\pm$ 0.02 & 0.505 $\pm$ 0.01 & \textbf{0.510} $\pm$ 0.02 &  0.329 $\pm$ 0.04 & 0.332 $\pm$ 0.01 \\
& GA2 + MA4  & 0.477 & 0.494 $\pm$ 0.02 & 0.489 $\pm$ 0.01 & 0.520 $\pm$ 0.01 & \textbf{0.526} $\pm$ 0.03 &  0.332 $\pm$ 0.01 &  0.336 $\pm$ 0.01 \\
        \midrule
\multicolumn{2}{c||}{Average AUROC} & 0.787 & 0.730 & 0.728 & 0.678 & 0.677 & 0.524 & 0.522  \\
\midrule
\multicolumn{2}{c||}{Average Rank} & 2.13 & 3.08 (**) & 2.78 (**) & 4.09 (***) & 3.74 (***) & 6.17 (***) &  6 (***)\\
        \bottomrule
    \end{tabular}
    \vspace{-0.1in}
\end{table*}

%% file: TABLES/auprc.tex
\begin{table*}
\centering
    \caption{Anomaly Detection Results for all methods across all datasets based on \textsc{AUPRC} (Area Under Precision-Recall Curve). For baselines,  average performance across hyperparameters along with the std. dev. is reported. \method outputs a unique ranking based on  a model selection criterion. Last row reports significance test results, where (**) and (***) denote that \method is significantly better than baselines w.r.t. the Wilcoxon Signed Rank Test at $p=0.05$ and $p=0.01$, respectively.}
    \vspace{-0.05in}
    \label{tab:ap_results}
    \setlength{\tabcolsep}{3pt}
    \begin{tabular}{c|c||c||c|c|c|c|c|c}
        \toprule
{Dataset}&
{Anomaly Type}
                    &  \method
                    & WL+BFS
                    &  WL+IR 
                    & G2V+BFS
                    & G2V+IR
                    & DOM.+BFS
                    & DOM.+IR \\
        \midrule
\multirow{6}{*}{SH}
& GA1  & \textbf{0.939}  &  0.470 $\pm$ 0.01 & 0.461 $\pm$ 0.01 &  0.270 $\pm$ 0.01 & 0.327 $\pm$ 0.01 & 0.327  $\pm$ 0.02 &  0.320 $\pm$ 0.01 \\
& GA2  &  \textbf{0.708} &  0.214 $\pm$ 0.02 &  0.217 $\pm$ 0.02 &  0.269 $\pm$ 0.04 & 0.252 $\pm$ 0.01 &  0.252 $\pm$ 0.01 &  0.256 $\pm$ 0.01 \\
& MA1  &  \textbf{0.280} &  0.125 $\pm$ 0.01 &  0.126 $\pm$ 0.01 &  0.118 $\pm$  0.01 &  0.125 $\pm$ 0.01 &  0.125 $\pm$ 0.01 &  0.126 $\pm$ 0.01 \\
& MA2  &   \textbf{0.599} &  0.125 $\pm$ 0.01 & 0.125 $\pm$ 0.01 &  0.120 $\pm$ 0.01 &  0.121 $\pm$ 0.01 &  0.121  $\pm$ 0.01 & 0.122 $\pm$ 0.01 \\
& GA1 + MA1  & \textbf{0.877} &  0.458$\pm$ 0.01&  0.450 $\pm$ 0.01 & 0.255 $\pm$ 0.09 & 0.246 $\pm$ 0.09 &  0.080 $\pm$ 0.01 &  0.077 $\pm$ 0.01 \\
& GA2 + MA1  &  \textbf{0.836} &  0.224 $\pm$ 0.02 &  0.225 $\pm$ 0.04 &0.286 $\pm$ 0.03 &  0.241 $\pm$ 0.01  &  0.097$\pm$ 0.01 &  0.093 $\pm$ 0.01  \\
        \hline
\multirow{6}{*}{KD}
& GA1  & \textbf{0.553} & 0.363 $\pm$ 0.02 &  0.347 $\pm$ 0.01 & 0.255 $\pm$ 0.09 & 0.246 $\pm$ 0.09 &  0.08 $\pm$ 0.01 &  0.080$\pm$ 0.01 \\
& GA2  & \textbf{0.430} & 0.284 $\pm$ 0.04 &  0.283 $\pm$ 0.04 & 0.234$\pm$ 0.014 & 0.240 $\pm$ 0.01 &   0.102 $\pm$ 0.01&   0.098 $\pm$ 0.01\\
& MA1  &  0.141 &  0.142 $\pm$ 0.01 & 0.136  $\pm$ 0.01 &  \textbf{0.148} $\pm$ 0.002 & 0.144 $\pm$ 0.01 &   0.106 $\pm$ 0.01 &  0.101 $\pm$ 0.01 \\
& MA2  & \textbf{0.342} &  0.117 $\pm$ 0.001 &  0.104 $\pm$ 0.01 &  0.162 $\pm$ 0.01 & 0.164 $\pm$ 0.01 &  0.089 $\pm$ 0.01 &  0.086 $\pm$ 0.01 \\
& GA1 + MA1  & \textbf{0.677} & 0.443 $\pm$ 0.01 &  0.424 $\pm$ 0.01 &  0.256 $\pm$ 0.12 & 0.247 $\pm$ 0.12 &   0.080 $\pm$ 0.01 &  0.077 $\pm$ 0.01 \\
& GA2 + MA1  & \textbf{0.351} & 0.261  $\pm$ 0.04 & 0.261 $\pm$ 0.04 & 0.237 $\pm$ 0.01 &  0.240 $\pm$ 0.01 &  0.097 $\pm$ 0.01 & 0.094 $\pm$ 0.01 \\
    \hline
\multirow{3}{*}{HW}
& GA1  & \textbf{0.866} & 0.446 $\pm$ 0.03 &  0.438 $\pm$ 0.03 & 0.455  $\pm$  0.10 & 0.451 $\pm$ 0.01&  0.314 $\pm$ 0.21 & 0.310 $\pm$ 0.21\\
& GA2  & \textbf{0.941} & 0.292 $\pm$  0.01&  0.280  $\pm$ 0.02 &  0.146 $\pm$  0.03 & 0.144 $\pm$ 0.03 &  0.106 $\pm$ 0.04 &  0.105 $\pm$ 0.04 \\
& MA2  & \textbf{0.815} &  0.165 $\pm$ 0.01 &  0.151 $\pm$  0.01 &  0.175 $\pm$ 0.01 & 0.168 $\pm$ 0.01&  0.114 $\pm$ 0.01 & 0.110 $\pm$ 0.01 \\
        \hline
\multirow{8}{*}{MobiNet}
& GA1  & 0.050 &  0.100 $\pm$ 0.01 &  \textbf{0.102} $\pm$ 0.01 & 0.046 $\pm$ 0.01& 0.046 $\pm$ 0.01 & 0.066$\pm$ 0.01& 0.065 $\pm$ 0.01 \\
& GA2  & 0.043 &  0.047 $\pm$ 0.01 &  0.047 $\pm$ 0.03 & 0.054 $\pm$ 0.01 &  0.054 $\pm$ 0.01 & \textbf{0.069} $\pm$ 0.01 & \textbf{ 0.069} $\pm$ 0.01\\
& MA3  & 0.040 &  0.054 $\pm$ 0.01  &  0.055 $\pm$ 0.01&  0.055 $\pm$ 0.01 &  0.057 $\pm$ 0.01&   \textbf{0.076} $\pm$ 0.01 &  \textbf{0.076}$\pm$ 0.01\\
& MA4  & 0.040 &  0.047 $\pm$ 0.01&  0.047 $\pm$ 0.01 &  0.051 $\pm$ 0.01 &  0.051 $\pm$ 0.01&  0.069 $\pm$ 0.01 & \textbf{0.067} $\pm$ 0.01 \\
& GA1 + MA3  & 0.052 & 0.165 $\pm$ 0.01 & \textbf{0.166} $\pm$ 0.01 & 0.043 $\pm$ 0.01 & 0.044 $\pm$ 0.01 & 0.067 $\pm$ 0.01 &  0.067 $\pm$ 0.01 \\
& GA1  + MA4 & 0.074 & 0.157  $\pm$  0.01 &  \textbf{0.158} $\pm$ 0.01 & 0.046 $\pm$ 0.01 & 0.046 $\pm$ 0.01  & 0.069 $\pm$ 0.01 & 0.068 $\pm$ 0.01 \\
& GA2 + MA3  & 0.052 & 0.044$\pm$ 0.01& 0.044 $\pm$ 0.01 & 0.054 $\pm$ 0.01 & 0.053 $\pm$ 0.01 &  \textbf{0.066} $\pm$ 0.01 &  0.065 $\pm$ 0.01 \\
& GA2 + MA4  & 0.041 &  0.046 $\pm$ 0.01  &  0.046 $\pm$ 0.01 & 0.055 $\pm$ 0.01 & 0.056 $\pm$ 0.01 & \textbf{0.066} $\pm$ 0.01 &   0.066 $\pm$ 0.01 \\
        \midrule
\multicolumn{2}{c||}{Average AUPRC} & 0.443 & 0.208 & 0.204 & 0.165 & 0.163 & 0.131 & 0.130  \\
\midrule
\multicolumn{2}{c||}{Average Rank} & 2.13 & 3.91 (***) & 4.26 (***) & 4.73 (***)& 3.95 (***)& 4.52 (***) & 4.48 (***) \\
        \bottomrule
    \end{tabular}
    \vspace{-0.1in}
\end{table*}

%% file: TABLES/ablation.tex
\begin{table*}[!t]
\centering
    \caption{Ablation Study Results - Comparing \method against its three variants: (i) \method without Metadata Fusion, (ii) \method without DeepSet \& (iii) \method with One-Class DeepSVDD loss (OCDL)}
    \vspace{-0.05in}
    \label{tab:ablation}
    \setlength{\tabcolsep}{3pt}
    \begin{tabular}{c|c|c|c||c|c||c|c||c|c}
        \toprule
\multirow{2}{*}{Dataset} &
\multirow{2}{*}{Anomaly Type}
                    &   \multicolumn{2}{c|}{\method}
                    &   \multicolumn{2}{c|}{\method \textbf{w/o Metadata}}
                    &   \multicolumn{2}{c|}{\method \textbf{w/o DeepSet}}
                    &   \multicolumn{2}{c}{\method \textbf{with OCDL}}
                    \\
        \cline{3-10}
                    & & AUROC & AUPRC & AUROC & AUPRC & AUROC & AUPRC & AUROC & AUPRC \\
        \midrule
\multicolumn{1}{c|}{\multirow{3}{*}{SH}}
& GA1  & \textbf{0.992} & \textbf{0.938} & 0.989 & 0.920 & 0.988  & 0.920 & 0.986 & 0.894\\
& GA2  &  \textbf{0.968} & \textbf{0.708} & 0.961 & 0.622 & 0.965 & 0.758 & 0.930 & 0.489\\
& MA2  &  \textbf{0.918} & \textbf{0.598} & 0.898 & 0.580 & 0.816 & 0.427 & 0.868 & 0.467 \\
        \midrule
        \multicolumn{1}{c|}{\multirow{2}{*}{MobiNet}}
& GA1  & 0.526 & 0.049 & \textbf{0.779} &  \textbf{0.179} & 0.588 & 0.060 & 0.577 & 0.061\\
& GA2  & 0.491 & 0.043 & \textbf{0.678}  & \textbf{0.079} & 0.475 & 0.042 & 0.472 & 0.045\\
        \bottomrule
    \end{tabular}
    \vspace{-0.15in}
\end{table*}

%% file: 05related.tex

Anomaly detection (AD) has an extensive literature mainly considering outliers in tabular or vector data  \cite{chandola2009anomaly,aggarwal2015outlier}, including the recently emerging deep neural network based approaches (see surveys \cite{chalapathy2019deep,journals/corr/abs-2009-11732,pang2021deep}). However, these do not apply to  AD for graphs with relational structure. 

 The majority of work on graph anomaly detection \cite{akoglu2015graph}, including the recent graph NN (GNN) based  techniques \cite{yu2015glad,conf/kdd/YuCAZCW18,conf/sdm/DingLBL19,ocgnn20} focus on node, edge, or subgraph anomalies within a \textit{single} graph, rather than  graph-level anomalies in a \textit{database}.


Different from these earlier work, we consider graph-level AD among a set of graphs within a database. 
There exist traditional encoding or compression-based techniques \cite{conf/kdd/NobleC03,eberle2009identifying,lee2021gawd,nguyen2023detecting} that aim to identify frequent structural motifs or graphlets that compress a graph database efficiently, and then flag those graphs with long encoding length as anomalous. 
Most recent work have shifted attention to employing deep learning and 
GNNs toward graph-based AD (for a recent survey, see 
\cite{ma2021comprehensive}).
The idea is to flatten each graph by leveraging their representation or embedding learning capability, and  train the GNN parameters end-to-end through various AD objectives such as one-class  \cite{zhao2023using,qiu2022raising}, mutual information-based \cite{zhang2022dual}, distributional distance \cite{zhao2022graph}, contrastive \cite{luo2022deep} as well as distillation losses \cite{ma2022deep}.
While these have made progress in graph-level anomaly detection, they do not  handle \textit{multi}-graphs, nor are they designed to admit multi-modal input such as graphs with meta-features as in our case.


Examples of prior work on multi-graphs address summarization
\cite{berberidis2022summarizing}, partitioning \cite{tang2009clustering,papalexakis2013more,kang2020multi}, as well as anomaly detection \cite{nguyen2023detecting,maruhashi2011multiaspectforensics}, however without considering additional meta-features. 
An earlier work on node-level (fake reviewers) AD in a single (reviewer-to-product) graph  has attempted to bridge node-level meta-features with graph data---by first using the meta-features to estimate node outlierness scores and then propagating those over the graph to capture guilt-by-association \cite{rayana2015collective}. 
Their method, however, does not generalize to graph database anomalies with graph-level meta-features. 

In summary our proposed \method, to our knowledge, is the first method for graph-level anomaly detection for directed node/edge-attributed multi-graphs with meta-features. It leverages ($i$) end-to-end \textit{multi}-graph embedding, ($ii$) \textit{joint} multi-modal representation learning and ($iii$) a \textit{multi-centroid} AD loss to effectively capture complexities in the input data.


%% file: 06conclusion.tex
In this work we addressed an anomaly detection problem that relates to one of the key challenges of big data mining, that is, data complexity. In particular, we considered a graph database consisting of  node- and edge-attributed directed multi-graphs with associated metadata, and proposed  a new  multi-modal anomaly detection approach called \method. 
In a unified neural network framework, \method first captures a set representation of the multi-edges, learns a graph-level embedding, fuses the graph and metadata in a joint embedding space, on which it finally employs an unsupervised anomaly loss based on a multi-centered data distribution. 
To our knowledge, \method is the first unified method that can tackle anomaly detection on complex data of this nature in an end-to-end fashion. Through extensive experiments on datasets from two real-world domains, namely accounting and urban mobility, we showed that \method significantly outperforms all two-stage baselines that handle graphs and metadata separately. We open-source \method's code for future research as well as  practical use on possibly other real-world domains.

%% file: 07appendix.tex
\subsection{Hyperparameter Configurations}
\label{sec:appendix}
\begin{table}[ht!]
\centering
\caption{Hyperparameter Configurations}
\label{tab:datasets}
\begin{tabular}{p{2cm} r}
\toprule
\textbf{Hyperparameter}  & \textbf{Configurations}\\ 
\midrule
\rowcolor{gray!30}
ADAMM &  \\
\# of centroids K & 1, 2, 4 \\
learning rate & 1e-4, 1e-3  \\
weight decay & 1e-5, 1e-4 \\
$\lambda_1$ & 0.1 \\
$\lambda_2$ & 0, 0.1 \\ 
\rowcolor{gray!30}
WL Kernel & \\
WL iterations & 1, 2, 4, 8, 16 \\
\rowcolor{gray!30}
G2V & \\
G2V iterations & 1, 2, 4, 8, 16 \\
\rowcolor{gray!30}
DOMINANT & \\
$\alpha$ & 0.4, 0.6 \\
\bottomrule
\end{tabular}
\end{table}

%% file: main.bbl
\begin{thebibliography}{10}
\providecommand{\url}[1]{#1}
\csname url@samestyle\endcsname
\providecommand{\newblock}{\relax}
\providecommand{\bibinfo}[2]{#2}
\providecommand{\BIBentrySTDinterwordspacing}{\spaceskip=0pt\relax}
\providecommand{\BIBentryALTinterwordstretchfactor}{4}
\providecommand{\BIBentryALTinterwordspacing}{\spaceskip=\fontdimen2\font plus
\BIBentryALTinterwordstretchfactor\fontdimen3\font minus \fontdimen4\font\relax}
\providecommand{\BIBforeignlanguage}[2]{{%
\expandafter\ifx\csname l@#1\endcsname\relax
\typeout{** WARNING: IEEEtran.bst: No hyphenation pattern has been}%
\typeout{** loaded for the language `#1'. Using the pattern for}%
\typeout{** the default language instead.}%
\else
\language=\csname l@#1\endcsname
\fi
#2}}
\providecommand{\BIBdecl}{\relax}
\BIBdecl

\bibitem{tax2004support}
D.~M. Tax and R.~P. Duin, ``Support vector data description,'' \emph{Machine learning}, vol.~54, pp. 45--66, 2004.

\bibitem{aggarwal2017introduction}
C.~C. Aggarwal and C.~C. Aggarwal, \emph{An introduction to outlier analysis}.\hskip 1em plus 0.5em minus 0.4em\relax Springer, 2017.

\bibitem{pang2021deep}
G.~Pang, C.~Shen, L.~Cao, and A.~V.~D. Hengel, ``Deep learning for anomaly detection: A review,'' \emph{ACM computing surveys (CSUR)}, vol.~54, no.~2, pp. 1--38, 2021.

\bibitem{gupta2013outlier}
M.~Gupta, J.~Gao, C.~C. Aggarwal, and J.~Han, ``Outlier detection for temporal data: A survey,'' \emph{IEEE Transactions on Knowledge and data Engineering}, vol.~26, no.~9, pp. 2250--2267, 2013.

\bibitem{choi2021deep}
K.~Choi, J.~Yi, C.~Park, and S.~Yoon, ``Deep learning for anomaly detection in time-series data: review, analysis, and guidelines,'' \emph{IEEE Access}, vol.~9, pp. 120\,043--120\,065, 2021.

\bibitem{akoglu2015graph}
L.~Akoglu, H.~Tong, and D.~Koutra, ``Graph based anomaly detection and description: a survey,'' \emph{Data mining and knowledge discovery}, vol.~29, pp. 626--688, 2015.

\bibitem{ma2021comprehensive}
X.~Ma, J.~Wu, S.~Xue, J.~Yang, C.~Zhou, Q.~Z. Sheng, H.~Xiong, and L.~Akoglu, ``A comprehensive survey on graph anomaly detection with deep learning,'' \emph{IEEE Trans. on Knowledge and Data Engineering}, 2021.

\bibitem{akoglu2021anomaly}
L.~Akoglu, ``Anomaly mining: Past, present and future,'' in \emph{International Conference on Information \& Knowledge Management}, 2021, pp. 1--2.

\bibitem{lee2021gawd}
M.-C. Lee, H.~T. Nguyen, D.~Berberidis, V.~S. Tseng, and L.~Akoglu, ``Gawd: graph anomaly detection in weighted directed graph databases,'' in \emph{IEEE/ACM ASONAM}, 2021, pp. 143--150.

\bibitem{nguyen2023detecting}
H.~T. Nguyen, P.~J. Liang, and L.~Akoglu, ``Detecting anomalous graphs in labeled multi-graph databases,'' \emph{ACM Transactions on Knowledge Discovery from Data}, vol.~17, no.~2, pp. 1--25, 2023.

\bibitem{zhao2023using}
L.~Zhao and L.~Akoglu, ``On using classification datasets to evaluate graph outlier detection: Peculiar observations and new insights,'' \emph{Big Data}, vol.~11, no.~3, pp. 151--180, 2023.

\bibitem{qiu2022raising}
C.~Qiu, M.~Kloft, S.~Mandt, and M.~Rudolph, ``Raising the bar in graph-level anomaly detection,'' \emph{arXiv preprint arXiv:2205.13845}, 2022.

\bibitem{zhao2022graph}
L.~Zhao, S.~Sawlani, A.~Srinivasan, and L.~Akoglu, ``Graph anomaly detection with unsupervised {GNN}s,'' \emph{Preprint arXiv:2210.09535}, 2022.

\bibitem{zhang2022dual}
G.~Zhang, Z.~Yang, J.~Wu, J.~Yang, S.~Xue, H.~Peng, J.~Su, C.~Zhou, Q.~Z. Sheng, L.~Akoglu \emph{et~al.}, ``Dual-discriminative graph neural network for imbalanced graph-level anomaly detection,'' \emph{Advances in Neural Information Processing Systems}, vol.~35, pp. 24\,144--24\,157, 2022.

\bibitem{deepset2017}
M.~Zaheer, S.~Kottur, S.~Ravanbakhsh, B.~Poczos, R.~R. Salakhutdinov, and A.~J. Smola, ``Deep sets,'' in \emph{NeurIPS}, 2017.

\bibitem{xu2018how}
K.~Xu, W.~Hu, J.~Leskovec, and S.~Jegelka, ``How powerful are graph neural networks?'' in \emph{ICLR}, 2019.

\bibitem{liang2023bookkeeping}
P.~J. Liang, ``Bookkeeping graphs: Computational theory and applications,'' \emph{Foundations and Trends{\textregistered} in Accounting}, vol.~17, no.~2, pp. 77--172, 2023.

\bibitem{schneider2013unravelling}
C.~M. Schneider, V.~Belik, T.~Couronn{\'e}, Z.~Smoreda, and M.~C. Gonz{\'a}lez, ``Unravelling daily human mobility motifs,'' \emph{Journal of The Royal Society Interface}, vol.~10, no.~84, p. 20130246, 2013.

\bibitem{ramesh2022hierarchical}
A.~Ramesh, P.~Dhariwal, A.~Nichol, C.~Chu, and M.~Chen, ``Hierarchical text-conditional image generation with clip latents,'' \emph{arXiv preprint arXiv:2204.06125}, 2022.

\bibitem{ruff2018deep}
L.~Ruff, R.~Vandermeulen, N.~Goernitz, L.~Deecke, S.~A. Siddiqui, A.~Binder, E.~M{\"u}ller, and M.~Kloft, ``Deep one-class classification,'' in \emph{ICML}, 2018, pp. 4393--4402.

\bibitem{das2012selecting}
A.~Das, A.~Dasgupta, and R.~Kumar, ``Selecting diverse features via spectral regularization,'' \emph{NeurIPS}, vol.~25, 2012.

\bibitem{shervashidze2011weisfeiler}
N.~Shervashidze, P.~Schweitzer, E.~J. Van~Leeuwen, K.~Mehlhorn, and K.~M. Borgwardt, ``Weisfeiler-lehman graph kernels.'' \emph{Journal of Machine Learning Research}, vol.~12, no.~9, 2011.

\bibitem{taylor2000support}
J.~S. Taylor and N.~Cristianini, ``Support vector machines and other kernel-based learning methods,'' \emph{Cambridge University}, 2000.

\bibitem{narayanan2017graph2vec}
A.~Narayanan, M.~Chandramohan, R.~Venkatesan, L.~Chen, Y.~Liu, and S.~Jaiswal, ``graph2vec: Learning distributed representations of graphs,'' \emph{arXiv preprint arXiv:1707.05005}, 2017.

\bibitem{ding2019deep}
K.~Ding, J.~Li, R.~Bhanushali, and H.~Liu, ``Deep anomaly detection on attributed networks,'' in \emph{Proceedings of the 2019 SIAM International Conference on Data Mining}.\hskip 1em plus 0.5em minus 0.4em\relax SIAM, 2019, pp. 594--602.

\bibitem{liu2008isolation}
F.~T. Liu, K.~M. Ting, and Z.-H. Zhou, ``Isolation forest,'' in \emph{ICDM}.\hskip 1em plus 0.5em minus 0.4em\relax IEEE, 2008, pp. 413--422.

\bibitem{emmott2015meta}
A.~Emmott, S.~Das, T.~Dietterich, A.~Fern, and W.-K. Wong, ``A meta-analysis of the anomaly detection problem,'' \emph{arXiv preprint arXiv:1503.01158}, 2015.

\bibitem{lazarevic2005feature}
A.~Lazarevic and V.~Kumar, ``Feature bagging for outlier detection,'' in \emph{ACM SIGKDD}, 2005, pp. 157--166.

\bibitem{chandola2009anomaly}
V.~Chandola, A.~Banerjee, and V.~Kumar, ``Anomaly detection: A survey,'' \emph{ACM computing surveys (CSUR)}, vol.~41, no.~3, pp. 1--58, 2009.

\bibitem{aggarwal2015outlier}
C.~C. Aggarwal, ``Outlier analysis,'' in \emph{Data mining}.\hskip 1em plus 0.5em minus 0.4em\relax Springer, 2015, pp. 237--263.

\bibitem{chalapathy2019deep}
R.~Chalapathy and S.~Chawla, ``Deep learning for anomaly detection: A survey,'' \emph{arXiv preprint arXiv:1901.03407}, 2019.

\bibitem{journals/corr/abs-2009-11732}
L.~Ruff, J.~R. Kauffmann, R.~A. Vandermeulen, G.~Montavon, W.~Samek, M.~Kloft, T.~G. Dietterich, and K.-R. M\"uller, ``A unifying review of deep and shallow anomaly detection.'' \emph{arXiv:2009.11732}, 2020.

\bibitem{yu2015glad}
R.~Yu, X.~He, and Y.~Liu, ``Glad: group anomaly detection in social media analysis,'' \emph{TKDD}, vol.~10, no.~2, pp. 1--22, 2015.

\bibitem{conf/kdd/YuCAZCW18}
W.~Yu, W.~Cheng, C.~C. Aggarwal, K.~Zhang, H.~Chen, and W.~Wang, ``Netwalk: A flexible deep embedding approach for anomaly detection in dynamic networks,'' in \emph{KDD}.\hskip 1em plus 0.5em minus 0.4em\relax ACM, 2018, pp. 2672--2681.

\bibitem{conf/sdm/DingLBL19}
K.~Ding, J.~Li, R.~Bhanushali, and H.~Liu, ``Deep anomaly detection on attributed networks,'' in \emph{SDM}.\hskip 1em plus 0.5em minus 0.4em\relax SIAM, 2019, pp. 594--602.

\bibitem{ocgnn20}
X.~Wang, Y.~Du, P.~Cui, and Y.~Yang, ``{OCGNN:} one-class classification with graph neural networks,'' \emph{CoRR}, vol. abs/2002.09594, 2020.

\bibitem{conf/kdd/NobleC03}
C.~C. Noble and D.~J. Cook, ``Graph-based anomaly detection.'' in \emph{KDD}.\hskip 1em plus 0.5em minus 0.4em\relax ACM, 2003, pp. 631--636.

\bibitem{eberle2009identifying}
W.~Eberle, L.~Holder, and D.~Cook, ``Identifying threats using graph-based anomaly detection,'' in \emph{Mach. Learn. in Cyber Trust}, 2009.

\bibitem{luo2022deep}
X.~Luo, J.~Wu, J.~Yang, S.~Xue, H.~Peng, C.~Zhou, H.~Chen, Z.~Li, and Q.~Z. Sheng, ``Deep graph level anomaly detection with contrastive learning,'' \emph{Scientific Reports}, vol.~12, no.~1, p. 19867, 2022.

\bibitem{ma2022deep}
R.~Ma, G.~Pang, L.~Chen, and A.~van~den Hengel, ``Deep graph-level anomaly detection by glocal knowledge distillation,'' in \emph{WSDM}.\hskip 1em plus 0.5em minus 0.4em\relax ACM, 2022, pp. 704--714.

\bibitem{berberidis2022summarizing}
D.~Berberidis, P.~J. Liang, and L.~Akoglu, ``Summarizing labeled multi-graphs,'' in \emph{Joint European Conference on Machine Learning and Knowledge Discovery in Databases}.\hskip 1em plus 0.5em minus 0.4em\relax Springer, 2022, pp. 53--68.

\bibitem{tang2009clustering}
W.~Tang, Z.~Lu, and I.~S. Dhillon, ``Clustering with multiple graphs,'' in \emph{ICDM}.\hskip 1em plus 0.5em minus 0.4em\relax IEEE, 2009, pp. 1016--1021.

\bibitem{papalexakis2013more}
E.~Papalexakis, L.~Akoglu, and D.~Ience, ``Do more views of a graph help? community detection and clustering in multi-graphs,'' in \emph{International Conference on Information Fusion}.\hskip 1em plus 0.5em minus 0.4em\relax IEEE, 2013, pp. 899--905.

\bibitem{kang2020multi}
Z.~Kang, G.~Shi, S.~Huang, W.~Chen, X.~Pu, J.~T. Zhou, and Z.~Xu, ``Multi-graph fusion for multi-view spectral clustering,'' \emph{Knowledge-Based Systems}, vol. 189, p. 105102, 2020.

\bibitem{maruhashi2011multiaspectforensics}
K.~Maruhashi, F.~Guo, and C.~Faloutsos, ``Multiaspectforensics: Pattern mining on large-scale heterogeneous networks with tensor analysis,'' in \emph{ASONAM}.\hskip 1em plus 0.5em minus 0.4em\relax IEEE/ACM, 2011, pp. 203--210.

\bibitem{rayana2015collective}
S.~Rayana and L.~Akoglu, ``Collective opinion spam detection: Bridging review networks and metadata,'' in \emph{SIGKDD}, 2015, pp. 985--994.

\end{thebibliography}
